\documentclass[runningheads]{llncs}
\usepackage[titlenumbered,ruled]{algorithm2e}
\usepackage{graphicx}
\usepackage{wrapfig}
\usepackage{makecell}
\usepackage{amssymb}

\usepackage{color}

\newcommand{\ddn}{{D\!D\!N}\xspace}
\newcommand{\ads}{{A\!D\!S}\xspace}
\newcommand{\erf}{{E\!R\!F}\xspace}
\newcommand{\lerf}{{L\!E\!R\!F}\xspace}
\newcommand{\cnn}{{C\!N\!N}\xspace}
\newcommand{\icc}{{I\!C\!C}\xspace}
\newcommand{\occ}{{O\!C\!C}\xspace}
\newcommand{\obj}{{O\!b\!j}\xspace}

\begin{document}
\title{Densely Decoded Networks with Adaptive Deep Supervision for Medical Image Segmentation}
%
%
\author{Suraj Mishra \and
Danny Z. Chen}
%
%
\institute{Dept. of Computer Science and Engineering, University of Notre Dame}
%
\maketitle              

\begin{abstract}
Medical image segmentation using deep neural networks has been highly successful. However, the effectiveness of these networks is often limited by inadequate dense prediction and inability to extract robust features. To achieve refined dense prediction, we propose densely decoded networks ($\ddn$), by selectively introducing \textit{`crutch'} network connections. Such \textit{`crutch'} connections in each upsampling stage of the network decoder (1) enhance target localization by incorporating high resolution features from the encoder, and (2) improve segmentation by facilitating multi-stage contextual information flow. Further, we present a training strategy based on adaptive deep supervision ($\ads$), which exploits and adapts specific attributes of input dataset, for robust feature extraction. In particular, $\ads$ strategically locates and deploys auxiliary supervision, by matching the average input object size with the layer-wise effective receptive fields ($\lerf$) of a network, resulting in a class of $\ddn$s. Such inclusion of \textit{'companion objective'} from a specific hidden layer, helps the model pay close attention to some distinct input-dependent features, which the network might otherwise \textit{`ignore'} during training. Our new networks and training strategy are validated on 4 diverse datasets of different modalities, demonstrating their effectiveness.
\end{abstract}

\section{Introduction}
In recent years, convolutional neural networks ($\cnn$) have been highly successful in medical image segmentation.
Every \textit{skeletonized} $\cnn$ for image segmentation task, can be viewed as a combination of an encoder and a decoder. While the encoder extracts features from the image domain, the decoder maps these extracted features into the mask domain. Such a mapping is done by assigning a class label to every pixel of the image, which is also known as localization. However, we observe that the network parameters may not be fully utilized in predicting dense pixels from coarse encoder features, which can lead to unrefined spatial localization and 
poor segmentation. Further, we believe that image domain specific information can be exploited for improved feature extraction. Enhancing segmentation accuracy by (1) distilling the dense prediction task and (2) consolidating feature extraction with an enhanced training strategy tailored specifically for biomedical image datasets are the major objectives of this work. 

Our first objective of distilling dense prediction has been a topic of interest.
Long et al.~\cite{fcn} proposed a ``fully convolutional network" along with \textit{`skip'} connections to avoid coarse output. Ronneberger et al.~\cite{unet} modified the model in \cite{fcn} by introducing stage wise upsampling to avoid larger strides. Each upsampling stage is supplemented by a corresponding stage in the encoder. A similar approach for 3D segmentation was proposed by Milletari et al.~\cite{vnet}. Despite their success, most of these architectures fail to segment the target objects accurately because of their inability to capture finer details. To achieve our first objective, we strategically divide the distilled dense prediction task into two sub-tasks: (1) enhanced localization and (2) finer detail generation. To tackle these two tasks, we propose the new densely decoded networks ($\ddn$s).

$\ddn$s are capable of generating refined dense prediction from coarse features. To accomplish this, $\ddn$s implement \textit{`crutch'} connections in every upsampling stage of the network. Such \textit{crutch} connections can be classified into two major categories: (1) \textit{incoming crutch connections} ($\icc$) and (2) \textit{outgoing crutch connections} ($\occ$).  While the commensurate stage inclusion using \textit{'crutch'} connections was proposed in \cite{unet}, $\icc$ (inspired by DenseNet \cite{densenet}) extends it by including all deeper stages too. Such inclusion contributes towards enhanced localization, as each upsampling stage is provided with feature maps in varying degrees of coarseness, which are combined to generate the output of the stage.  While $\icc$ is included for enhanced localization, $\occ$ is directed towards finer detail generation. $\occ s$ (inspired by \cite{retinanet,fpn}) are extracted from each upsampling stage, and are merged to generate the final output. A somewhat similar approach to $\occ$ was proposed in \cite{cumednet}. We modify it by generating $\occ$ from the stage-wise decoder instead of the encoder and including $\occ$ from every stage.

Robust discriminativeness among features is a vital attribute for any feature extractor. In addition to accounting for the issue of vanishing gradient, using auxiliary classifiers to perform deep supervision has been effective in ensuring discriminativeness of features in hidden layers \cite{lee}. Hence, we leverage deep supervision for our second objective of robust feature extraction. Deep supervision can be viewed as a feedback generated by a local output (from a hidden layer) by providing companion objective with auxiliary classifiers \cite{lee}. The combined error is back-propagated, influencing the hidden layer update process to favor highly discriminative feature maps. The strategy of how to introduce deep supervision into a network architecture has been an active research topic \cite{szegedy,lee,cumednet,densevoxnet}, but has not been explored comprehensively. We extend the idea of deep supervision to biomedical image segmentation by proposing adaptive deep supervision ($\ads$). $\ads$ introduces deep supervision from the specific layer in the encoder that has the preeminent contribution towards discriminative feature extraction. $\ads$ exploits the relative \textit{`stable'} nature of biomedical image data to extract dataset specific information (e.g., the average object size), and matches this information with the receptive field of the network to determine the preeminent layer adaptively for different specific datasets. This results in a class of `adaptive' networks capable of extracting improved robust features.

We extensively validate each component of our proposed framework using four datasets of differing modalities (Ultrasound, 3D MRI, RGB, and Grayscale). Our experiments show that combining $\ddn$ with $\ads$ generates a class of networks which are capable of being trained end-to-end while achieving better segmentation accuracy over state-of-the-art methods.

\section{Method}

\begin{figure}[t]
\centering
\includegraphics[width=11.25cm,height=3.5cm]{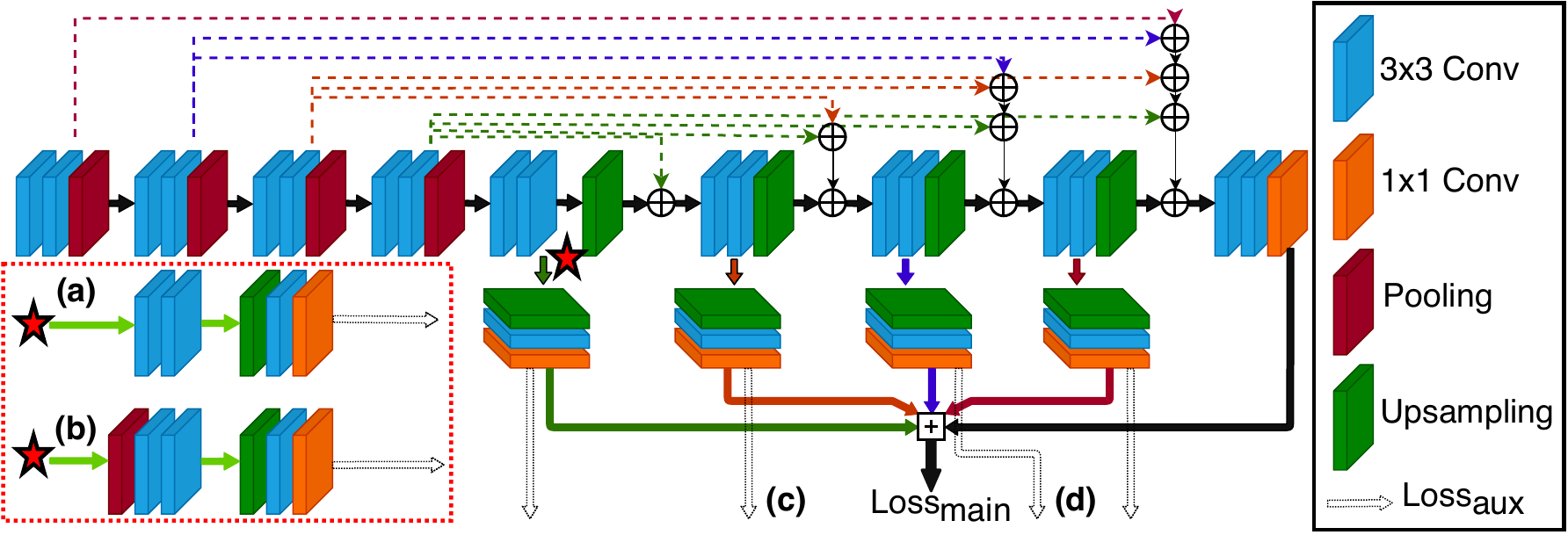}
\caption{Our proposed DDN architecture. ICC is shown as dashed colored arrows while OCC is thick colored arrows. ADS for (a) melanoma, (b) wing disc (red dotted box for the Obj $>$ LERF$_N$ cases), (c) lymph node, (d) BraTS are shown as dotted colorless arrows. Each learnable layer (except 1x1 conv) is followed by batch-norm and ReLU activation. Upsampling layers in ICC (similar as in OCC) are not shown here.} 
\label{fig:ddn}
\end{figure}

\subsection{Densely Decoded Networks (DDN)}
\label{ssec:net}
We propose $\ddn$, with a U-Net \cite{unet} type encoder-decoder architecture, for biomedical image segmentation. Fig.~\ref{fig:ddn} shows our proposed network 
along with \textit{`crutch'} connections. We first discuss $\icc$, followed by the details of $\occ$.

\textbf{ICC}: Consider a $\cnn$, with $N$ stacked conv-layers in its encoder, arranged in $M$ stages. Each stage represents a set of conv-layers extracting features with a fixed feature map size (i.e., in-between two down-sampling operations). Assuming a U-Net \cite{unet} type \textit{skeleton} network, there is a corresponding upsampling stage for each downsampling stage. Denoting $y_m$ as the output of the $m^{th}$ upsampler $U_m$, we can state $y_m = U_m(x_m)$, where $x_m$ is the input to $U_m$. 
In $\ddn$, each upsampling stage takes $\icc$s from the commensurate and all the deeper stages in the encoder. Hence, $y_m = U_m(x_m \oplus E_m \oplus E_{m+1} \oplus \cdots \oplus E_M)$, where $E_i$ is the output of the encoder stage $i$ (or $\icc$), $i=m,\ldots,M$ and $\oplus$ is the concatenation operation. Note that U-Net can be viewed as a special case of $DDN$ since for U-Net, $y_m = U_m(x_m \oplus E_m)$, with $m \in M$. Each $\icc$ acts as crutches to the main input to any upsampling stage (the main input $x_m$, black arrows in Fig.~\ref{fig:ddn}), resulting in enhanced localization. Intuitively, inclusion of feature maps of varying coarseness degrees to generate the upsampling output is the key here.

\textbf{OCC}: Output of a $\cnn$ can be stated as $y_0 = U_1(x_1)$, where $U_1$ is the last upsampler with input $x_1$, generating the output $y_0$ (ignoring conv layers between $U_1$ and the output). In $\ddn$, the final output is a combination of outputs (pixel-wise feature map addition, shown as $\boxplus$ in Fig.~\ref{fig:ddn}) from all the upsampling stages, i.e., $y_0 = U_1(x_1) + \cdots + U_M(x_M)$ or $\occ$s (the vanilla case, without considering $\icc$s). $\occ$s can be viewed as crutches for the final output. Inclusion of upsampled features from various feature map sizes facilitates multi-stage contextual information flow, resulting in finer detail generation. Merging both $\icc$s and $\occ$s, the final output of $\ddn$ can be reported as $y_0 = U_1(x_1 \oplus \cdots \oplus E_M) + \cdots + U_M(x_m)$, where $x_m = E_M$. Intuitively, $DDN$ has a more general architecture with many current architectures as its special cases. 

\begin{figure}[t]
\includegraphics[width=11.5cm,height=3.25cm]{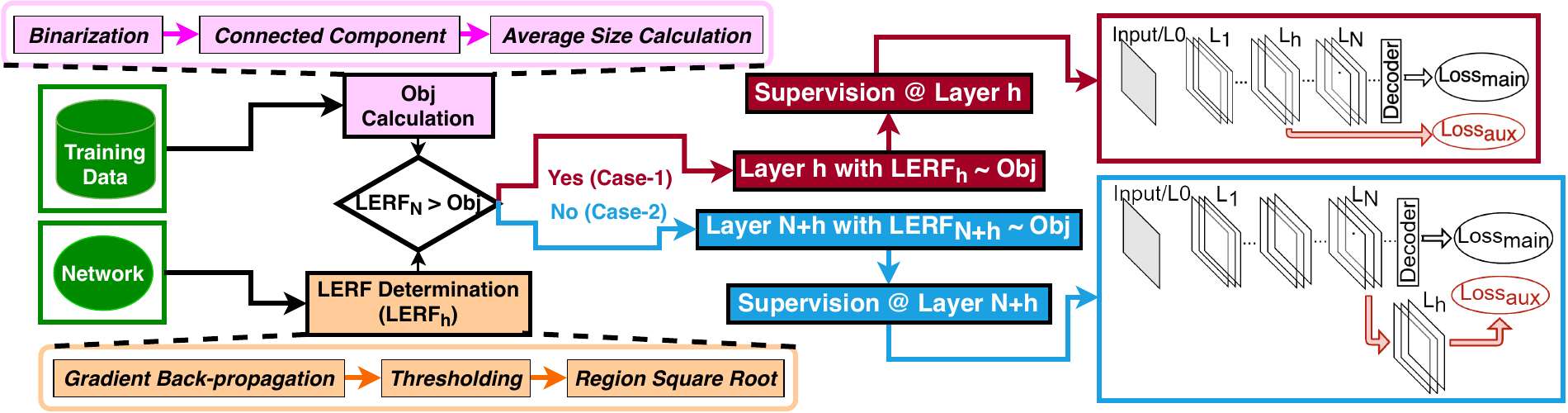}
\centering
\caption{Our proposed model for ADS. Based on the (LERF, Obj) matching (i.e., inputs, shown in green), ADS suggests the location of Loss$_{aux}$, shown in red and blue boxes.} 
\label{fig:ads}
\end{figure}

\subsection{Adaptive Deep Supervision (ADS)}
\label{ssec:ads}
For robust feature extraction, we propose $\ads$, which deals with the receptive field of a network, to introduce auxiliary classifiers. Convolutional layers in $\cnn$s use only a portion of the input, which contributes towards the output generation and is known as the {\it receptive field} ($RF$) of a layer \cite{Lecun}. For a $\cnn$ performing segmentation, $RF$ of the network is regarded as the region of input influencing the output of the encoder. Luo et al.~\cite{erf} showed that this region of influence is smaller (called the effective receptive field or $\erf$) with unresponsive border. Intuitively, convolutional filters extract improved features when the \textit{target object size} matches the filters' $\erf$ and the object lies at the $\erf$ center~\cite{mishra-vessel,mishra_isbi21,ds-tmi,mishra-skin}. 

We propose, for $\cnn$s performing biomedical image segmentation, using deep supervision from a specific hidden layer (the {\it preeminent layer}) to introduce an companion objective (or regularization), which forces the network not to \textit{ignore} some specific input-dependent features. Our method is based on the intuition that smaller objects are lost in the network's sub-sampling operations, while robust global features for larger objects are never extracted due to a smaller network $\erf$ or \textit{field of view}. Hence, we consider the preeminent layer determination as a matching problem, whose objective is to determine the network layer whose $\erf$ matches as well as possible with the target object size ($\obj$) of the input dataset. We modify the method in \cite{erf} by proposing layer-wise $\erf$ ($\lerf$) and use it instead of $ERF$ to perform matching (explained below). The framework for $\ads$ is highlighted in Fig.~\ref{fig:ads}. The major components of $\ads$ are: (1) $\obj$ calculation, (2) $\lerf$ determination, and (3) $\ads$ implementation. We provide detailed explanations of each components in the following sections. 

\textbf{Obj Calculation}: Unlike natural scene images, objects in biomedical images are relatively `stable' (e.g., a dataset is for a specific disease or type of organs, captured in a specific modality). We exploit this property by estimating a representative object size ($\obj$) for a specific dataset using the mean object size with an acceptable level of accuracy. We use a connected-component based method for object size calculation on the training set of labelled
images $G_i$, $i =1,2,\ldots,g$, where each image has $J_i$ objects. Binarized labelled images ($\mathbb{R}^C \to \mathbb{R}^{0,1}$) are used to determine the pixel area of each target object ($T_i^j$ in $G_i$, i.e, the $j^{th}$ object in the $i^{th}$ image), irrespective of the target classes $C$, by treating each target object as a connected-component. The square root of the pixel area is considered as the size of an object ($o_i^j = \sqrt{PixelArea(T_i^j)}$). The mean value of these object sizes is taken as the average object size for $G_i$. The dataset characteristic object size is the mean value over all the $g$ images ($\obj = \sum_{i}(\sum_{j} o_i^j / J_i) / g$).

\textbf{LERF Determination}: We adopt a layer-wise approach for $\erf$ calculation (i.e., $\lerf$), since our objective is to match $\obj$ with the $\erf$ of a specific layer. We follow a partial derivative based approach to measure the influence of the input region, over an output node (a unit feature map, i.e., a pixel) of a conv layer \cite{erf}. For a CNN with $N$ conv layers in its encoder, $\lerf_h$ is associated with layer $h$ ($h = 1,2,\ldots,N$). With an input node $p_{i,j}$, our objective is to determine $\frac{\partial q_{i_t,j_t}}{\partial p_{i,j}}$, where $q_{i_t,j_t}$ is the target node at the output of the conv layer for which the influence measurement is being performed. In a $\cnn$, error gradients are calculated during back-propagation (i.e., $\frac{\partial L}{\partial p_{i,j}}$, for a loss function $L$). Hence, by assigning the target node gradient a non-zero value ($\frac{\partial L}{\partial q_{i_t,j_t}} = 1$ and $\frac{\partial L}{\partial p_{i,j}} = 0$, for $i_t \neq i$ and $j_t \neq j$) and using the chain rule, the region of influence can be determined (i.e., all non-zero gradients at the input). This region has a near-\textit{Gaussian} distribution \cite{erf} with an insignificant border area. Therefore, the square root of the thresholded region is assigned as the $\lerf$ of the conv layer.

\textbf{ADS Implementation}: To implement $\ads$, we use $\obj$ and $\lerf$, to determine $L_{target}$, i.e., the conv layer with preeminent contribution towards feature extraction. In $\ads$ implementation, two possible cases must be tackled separately. \textbf{Case-1:} When $\lerf_N$ (i.e., $\erf$ of the network) is $\geq \obj$, $L_{target}$ is selected as the layer $h*$ where, $|\obj - \lerf_{h^*}| \leq |\obj - \lerf_h|; \forall ~ 1 \leq h \leq N$. Deep supervision is then implemented from $L_{target}$ to inculcate the hidden features extracted as companion objective for discriminative feature learning. Our experiments revealed that deep supervision using $OCC$ of the decoder (shown in Fig.~\ref{fig:ddn}) that has direct link with $L_{target}$ generates more accurate results. Further, such inclusion restricts additional network weights since dense features are already upsampled and hence preferred. When no direct connection between the decoder and $L_{target}$, the next conv layer $L_{h^t}$ is selected with $h^t > target$. \textbf{Case-2:} When $\lerf_N < \obj$, we propose stacking $h^*$ additional conv layers on the $Loss_{aux}$ branch extracted from $L_N$ (shown as $\star$ in Fig.~\ref{fig:ddn}), such that $|\obj - \lerf_{h^*}| \leq |\obj - \lerf_h|; \forall ~ 1 \leq h \leq N+h^*$. This stacking provides a network path with a larger $\erf$ compared to the base network, resulting in closer matching to $\obj$ and hence better global feature extraction.

Inclusion of $\ads$ also changes the total loss ($Loss_{total}$, for back-propagation) into the sum of the main loss ($Loss_{main}$) and auxiliary loss ($Loss_{aux}$), i.e., $Loss_{total} = Loss_{main} + Loss_{aux} + L_2$-$Regularization$. We do not use any multiplier to reduce the effect of $\ads$, instead letting the network learn by itself. We investigated cross-entropy, Dice, and Jaccard based $Loss_{main}$ and $Loss_{aux}$ calculations, and provide experimental details in the next section.

\section{Experiments and Results}
In this section, we first provide details of datasets used to validate our approach. Then we provide the general experimental setup to implement $\ddn$ for each dataset. $\ads$ implementation for individual cases are detailed after that. Finally, we discuss results obtained along with ablation study to corroborate our findings.

\textbf{(1) Melanoma segmentation in RGB images:}  The ISIC 2017 skin lesion dataset \cite{isic} contains 2000 training, 150 validation, and 600 test RGB images for melanoma segmentation. Observing the smaller validation set, we merge the training and validation sets and randomly select 20\% of the merged set for validation as done in \cite{me_cvpr_data}. Images are resized to $320 \times 320$ following \cite{mel_320}. \textbf{(2) Brain tumor segmentation in 3D MR images:} BraTS 2017 training dataset \cite{brats1,brats2,brats3} contains 285 cases (210 HGG and 75 LGG) of volumetric data in four modalities (T1, T1CE, T2, and FLAIR). Each volume is of size $240 \times 240 \times 155$ and was annotated into 3 tumor regions. For our experiments, we consider the more difficult LGG cases. As pre-processing, we normalize each image by subtracting mean and dividing variance. Following \cite{brats_flair}, we only consider the FLAIR modality and extract transverse slices (XY-plane) as training images. 3 classes of tumor regions are combined to generate the whole tumor region and used as ground truth for training. \textbf{(3) Lymph node segmentation in ultrasound images:} Lymph node dataset contains ultrasound images of the lymph node areas of 237 patients. Following \cite{yizhe}, we use 137 images for training (20\% for validation) and the rest for testing, assuring no identity overlap. Each image is resized to $224 \times 224$. \textbf{(4) Wing disc segmentation in grayscale images:} Wing disc pouches of fruit flies are used to explore organ development \cite{peixian}. 996 grayscale wing disc pouch images are investigated by using 889 images for training (20\% for validation) and 107 images for testing. Following \cite{peixian,suraj}, we use images of size $512 \times 512$ for the experiments.

\begin{figure}[t]
\centering
\includegraphics[width=11.5cm,height=7.3cm]{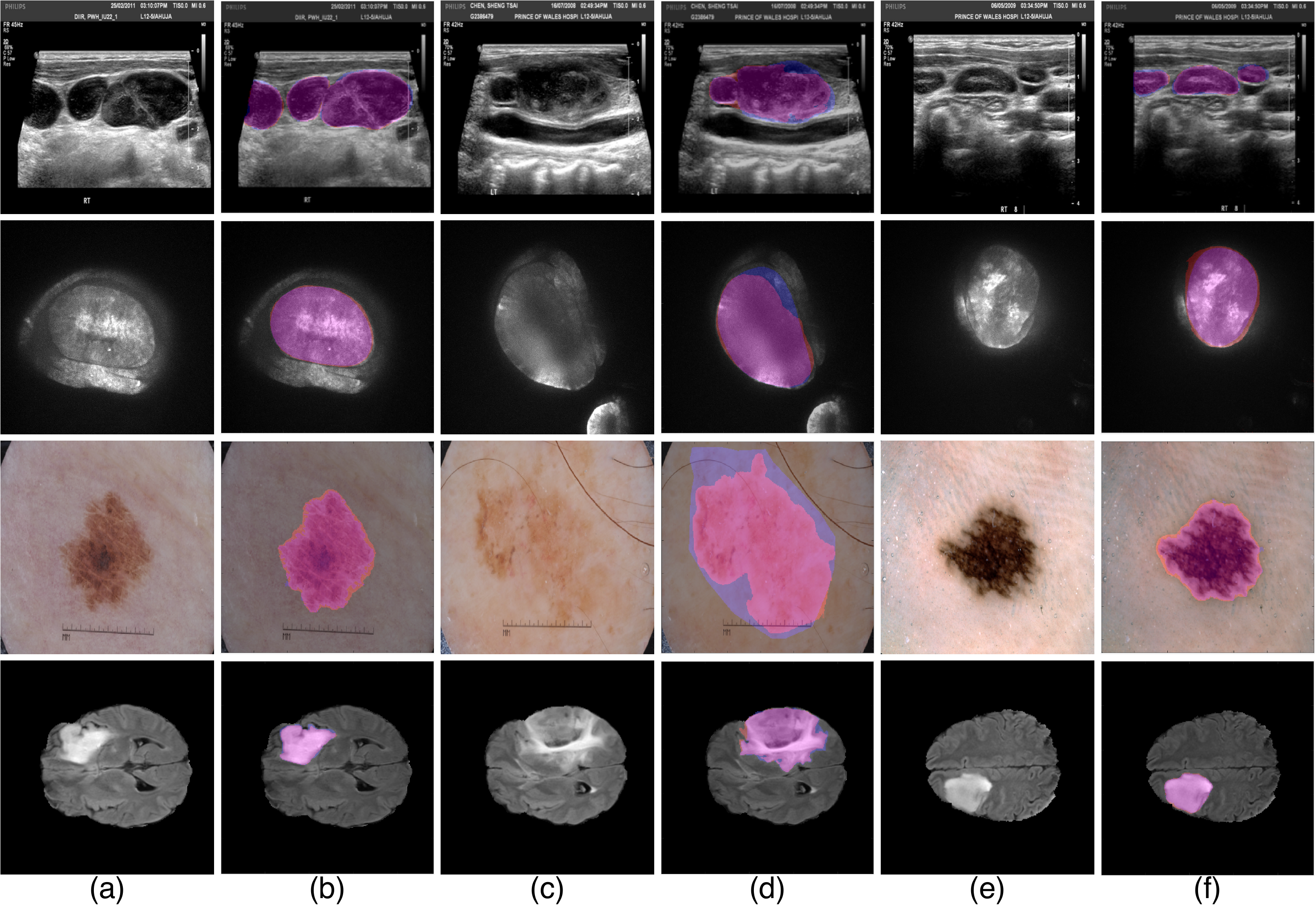}
\caption{Example segmentation results for (row-1) lymph node, (row-2) wing disc, (row-3) melanoma, and (row-4) BraTS dataset. (a), (c), and (e) show example test images. (b), (d), and (f) are the ground truth (blue) and segmentation output (red) obtained using DDN + ADS. Magenta (red + blue) highlights true positive regions.}  
\label{fig:results}
\end{figure}

\textbf{DDN Setup:} Our proposed $\ddn$ architecture (shown in Fig.~\ref{fig:ddn}) has a similar number of filter arrangements (or channels) in each layer as U-Net \cite{unet}. Initial experiments with different arrangements of dropout layers \cite{dropout} along with different dropout rates reveal that using a single dropout layer at the end of the encoder with a dropout rate of 50\% performs the best. Random flipping and rotation are used for data augmentation. We use a standard back-propagation implementing \textit{ADAM} with a fixed learning rate of 0.00002 ($\beta_1 = 0.9, \beta_2 = 0.999, \epsilon = 1\mathrm{e}{-8}$). Experiments were performed on NVIDIA-TITAN and Tesla P100 GPUs, using the PyTorch framework for a number of epochs (melanoma = 3000, BraTS = 500, lymph = 20000, wing disc = 5000). The batch size for each case is selected as the maximum size permissible by the GPU. We use similar network initialization as in He et al.~\cite{he_init}. We experimented with Dice and Jaccard loss for melanoma segmentation following \cite{mel_winner} but did not observe any significant change in accuracy compared to cross-entropy loss. While cross-entropy is used as $Loss_{main}$ for melanoma, lymph node, and wing disc segmentation, Dice loss is used for BraTS segmentation following \cite{brats_flair}. \textbf{ADS Setup:} Analyzing the \textit{Gaussian} distribution of $RF$ \cite{erf}, a threshold of 2 standard deviation ($95.4\%$) is used for $\lerf$ extraction. To neutralize variations in $\lerf$ calculations caused by network nonlinearity, the mean over 20 iterations is used. Following the implementation details provided in Sec.~\ref{ssec:ads}, $\obj$ for resized images is determined (melanoma $\sim$ 118, BraTS $\sim$ 16, lymph node $\sim$ 51, wing disc $\sim$ 189). Using $\lerf$ values of $\ddn$ layers, $\ads$ locations are decided and are shown in Fig.~\ref{fig:ddn}. We find that stacking conv-layers with 64 filter channels (instead of 1024 or 512 for Case-2) on $\ads$ results in better accuracy by avoiding overfitting and also results in a lesser number of network weights. We use the same $Loss_{aux}$ as $Loss_{main}$ for every $\ads$ case.

\begin{table}[tb]
\caption{Segmentation results.}
\scriptsize
\begin{center}
\scalebox{0.92}{
\begin{tabular}{ c c  c  c  c || c c  c  c  c }
\hline
\multicolumn{1}{c|}{} & \multicolumn{4}{c||}{Lymph Node} & \multicolumn{1}{c|}{} & \multicolumn{2}{c}{Wing Disc} \\
\hline
\rule{0pt}{8pt} Method & JAC & PR & RE & F1 & Method & meanIU & F1 \\ 
\Xhline{3\arrayrulewidth}
U-Net~\cite{unet} & 0.661 & 0.834 & 0.761 & 0.796 & U-Net~\cite{unet} & 0.948 & 0.954 \\ 
CUMedNet~\cite{cumednet} & 0.759 & 0.847 & 0.880 & 0.863 & CUMedNet~\cite{cumednet} & 0.939 & 0.945\\
Zhang et al.~\cite{yizhe} & 0.810 & 0.901 & 0.889 & 0.895 & Liang et al.~\cite{peixian} & 0.962 & 0.968\\
$DDN$ & 0.804 & 0.880 & 0.881 & 0.886 & $DDN$ & 0.968 & 0.972\\
$DDN+ADS$ & \textbf{0.827} & \textbf{0.903} & \textbf{0.893} & \textbf{0.908} & $DDN+ADS$ & \textbf{0.972} & \textbf{0.975}\\
\hline
\end{tabular}}

\scalebox{0.92}{
\begin{tabular}{ c c  c  c  c || c c  c  c  c }
\hline
\multicolumn{1}{c|}{} & \multicolumn{4}{c||}{Melanoma} & \multicolumn{1}{c|}{} & \multicolumn{2}{c}{BraTS} \\
\hline
\rule{0pt}{8pt} Method & JAC & SE & SP & Dice & Method & Dice & SE & SP \\ 
\Xhline{3\arrayrulewidth}

Li et al.~\cite{mel_320} & 0.753 & \textbf{0.855} & 0.974 & 0.839 & FCN~\cite{fcn} & 0.825$\pm$0.002 & 0.818 & 0.998 \\ 
Yuan et al.~\cite{mel_winner} & 0.765 & 0.825 & 0.975 & 0.849 & U-Net~\cite{unet} & 0.844$\pm$0.003 & 0.834 & 0.998\\
Bisla et al.~\cite{bisla} & 0.770 & - & - & - & Dong et al.~\cite{brats_flair} & 0.840 & - & -\\
$DDN$ & 0.772 & 0.854 & 0.966 & 0.865 & $DDN$ & 0.851$\pm$0.003 & 0.838 & 0.999\\
$DDN+ADS$ & \textbf{0.784} & 0.852 & \textbf{0.976} & \textbf{0.872} & $DDN+ADS$ & 0.864$\pm$0.002 & 0.840 & 0.999\\
\hline
\end{tabular}}
\end{center}
\label{tab:eval}
\end{table}

\begin{figure}[t]
\centering
\includegraphics[width=11.95cm,height=1.94cm]{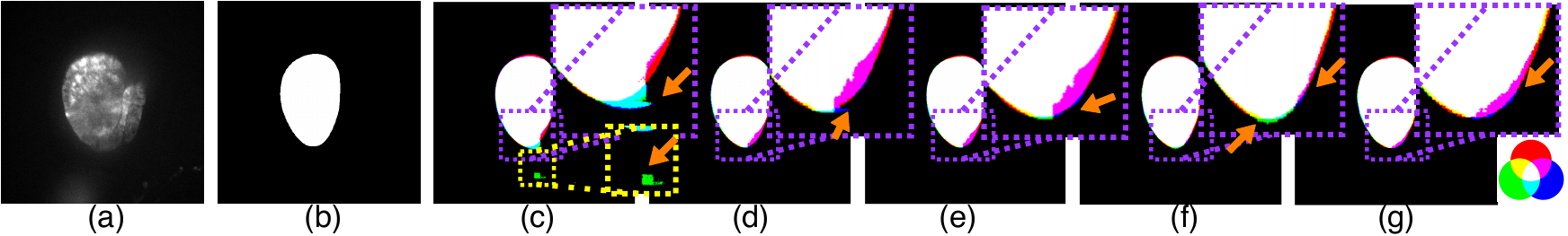}
\caption{(a) A wing-disc test image. (b) Ground truth (GT). (c) GT in red, FCN (output) in green, and FCN+ADS in blue. (d) GT in red, DDN-ICC in green, and DDN-ICC+ADS in blue. (e) GT in red, DDN-ICC in green, and DDN+ADS in blue. (f) GT in red, DDN-OCC+ADS in green, and DDN+ADS in blue. (g) GT in red, DDN+ADS$^*$ (ADS from (a) in Fig.~\ref{fig:ddn}) in green, DDN+ADS in blue ('-': without; '+': with).}  
\label{fig:abl}
\end{figure}

\textbf{Evaluation and Ablation Study:} Table~\ref{tab:eval} summarizes quantitative results obtained for all the four datasets, while Fig.~\ref{fig:results} shows qualitative results for example test cases. Similar post-processing and evaluation metrics as the existing state-of-the-art are used for each case. For lymph node, $\ddn+\ads$ achieves the state-of-the-art accuracy in all the 4 metrics while reducing the number of network parameters by  $\sim 50\%$. $\ddn$ itself attains the state-of-the-art accuracy for wing disc segmentation. Implementing $\ads$ on top of $DDN$ further improves the performance. For the melanoma dataset, $\ddn+\ads$ shows more impressive accuracy gain with $\sim 3\%$ increase in Dice coefficient. Following \cite{mel_winner}, we explored different inputs like stacking color channels (RGB+HSV, RGB+HSV+L), but did not observe any significant change in accuracy. Similar improved accuracy is also observed on BraTS data. Isensee et al. \cite{isensee} has reported higher Dice (0.895) compared to $\ddn + \ads$, however, using both HGG and LGG data. Further, we believe $\ddn$ implementing 2D filters cannot extract inter-slice context from 3D volumes. \textbf{Ablation Study:} Benefit of $\ads$ is provided in the last two rows of the respective dataset in Table~\ref{tab:eval} and is also shown in Fig.~\ref{fig:abl}(c), (d) and (e) (magenta area highlighting increase in true positive cases). We also performed similar studies on U-Net~\cite{unet} and FCN~\cite{fcn} type base networks (Fig.~\ref{fig:abl}(c)). Similarly contributions of $\icc$ and $\occ$ are studied and one example case is provided in Fig.~\ref{fig:abl}(d), (e), and (f). Further experiments on $\ads$ location selection (Fig.~\ref{fig:abl}(g)), corroborates our framework. More results can be found in Suppl. Materials.

\section{Conclusions}
In this work, we introduced densely decoded networks ($\ddn$) with \textit{`crutch'} connections, for enhanced localization and segmentation output generation. Further, we proposed adaptive deep supervision ($\ads$) that uses input data to guide the auxiliary supervision placement, demonstrating its ability in robust feature extraction. We exhibited the effectiveness of $\ddn$ and $\ads$ by conducting experiments on four datasets of different modalities. 

%
%
%
{
\bibliographystyle{splncs04}
\bibliography{refs}
}

\end{document}


\title{Supplementary Materials}
\author{Suraj Mishra \and
Danny Z. Chen}
\institute{}
\maketitle  

\begin{figure}
\centering
\includegraphics[width=12.2cm,height=4.4cm]{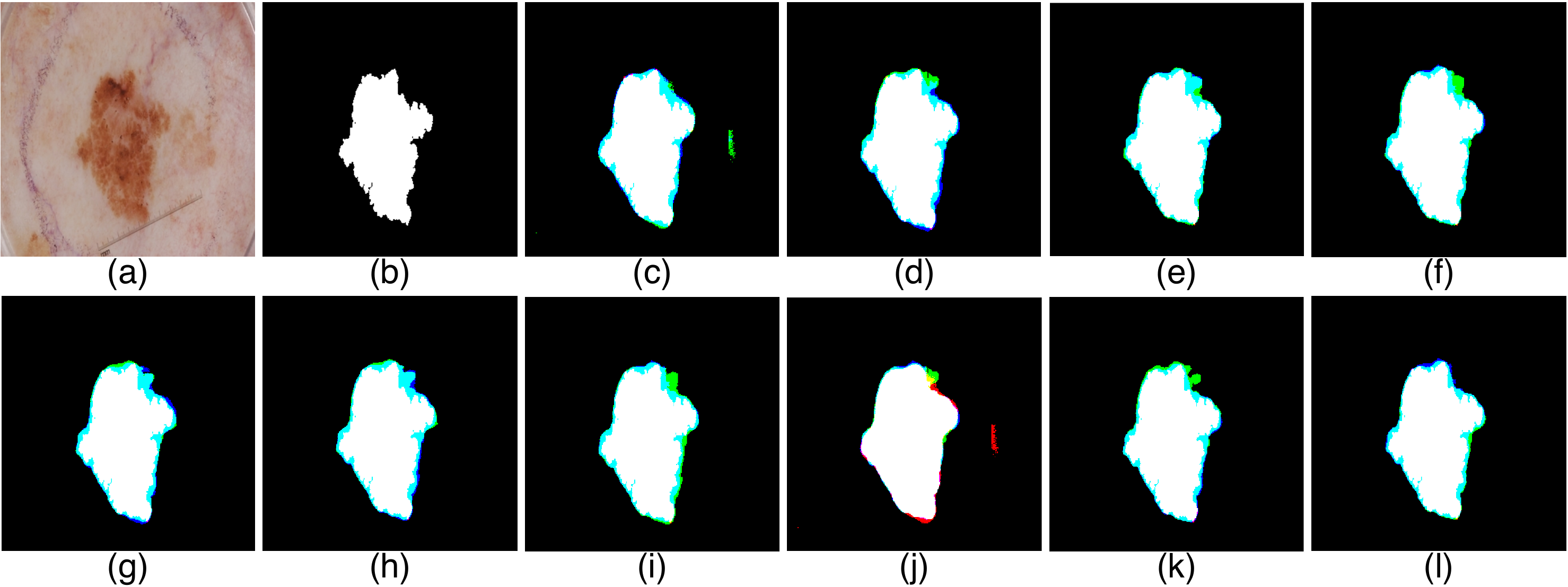}
\caption{(a) A melanoma image. (b) Ground truth (GT). (c) GT in red, FCN (output) in green, and FCN+ADS in blue. (d) GT in red, U-Net in green, and U-Net with RGB+HSV+L as input and trained with Jaccard loss in blue. (e) GT in red, U-Net in green, and U-Net+ADS in blue. (f) GT in red, U-Net+ADS in green, and DDN+ADS in blue. (g) GT in red, DDN-ICC in green, and DDN-ICC trained with Jaccard loss in blue. (h) GT in red, DDN-ICC in green, and DDN-ICC+ADS in blue. (i) GT in red, DDN-ICC+ADS in green, and DDN+ADS in blue. (j) FCN in red, U-Net in green, and DDN+ADS in blue. (k) GT in red, DDN in green, and DDN+ADS in blue. (l) GT in red, DDN+ADS without dropout in green, and DDN+ADS with dropout in blue.}  

\label{fig:results_me}
\end{figure}

\begin{figure}
\centering
\includegraphics[width=12.2cm,height=2.4cm]{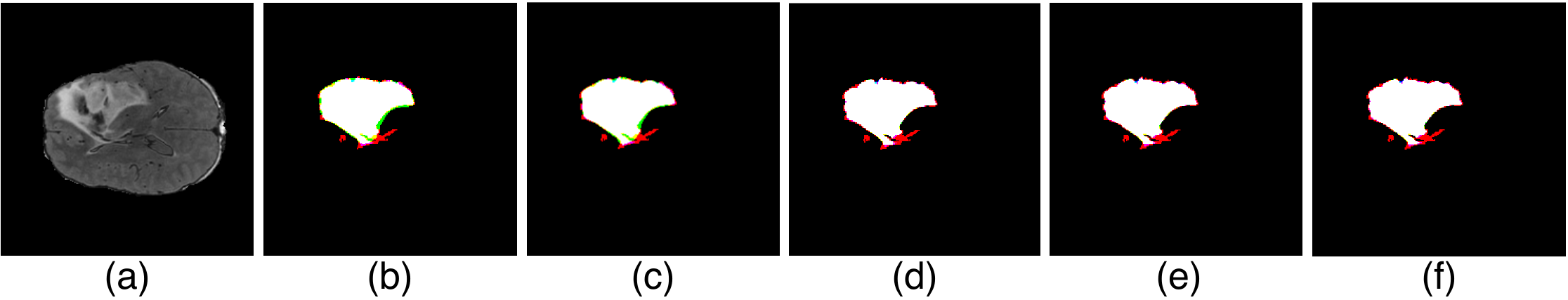}
\caption{Improved boundary segmentation. (a) A raw BraTS image. (b) Ground truth (GT) in red, FCN (output) in green, and DDN+ADS in blue. (c) GT in red, FCN+ADS in green, and U-Net+ADS in blue. (d) GT in red, U-Net in green, and U-Net+ADS in blue. (e) GT in red, U-Net+ADS in green, and DDN+ADS in blue. (f) GT in red, DDN in green, and DDN+ADS in blue.}  

\label{fig:results_wd}
\end{figure}

\begin{table}[tb]
\caption{Melanoma segmentation results. In U-Net$^*$ and  U-Net$^{**}$, Jaccarrd loss is used for training. In U-Net$^{**}$, stacked color channels (RGB+HSV+L) as input are used. In all the other cases, RGB input is used with cross-entropy loss. For U-Net+ADS$^{*}$, 1024 channels are used for conv layers on the ADS branch, while in all the other ADS cases 64 channels are used for conv layers.}
\scriptsize
\begin{center}
\scalebox{1.15}{
\begin{tabular}{ c | c | c | c | c | c | c | c | c | c | c | c}
\hline
\rule{0pt}{8pt} \rot{Method} & \rot{FCN~\cite{fcn}} & \rot{FCN+ADS} & \rot{U-Net~\cite{unet}} & \rot{U-Net$^{*}$} & \rot{U-Net$^{**}$} & \rot{U-Net+ADS} & \rot{U-Net+ADS$^{*}$} & \rot{DDN-ICC} & \rot{DDN-ICC+ADS} & \rot{DDN} & \rot{DDN+ADS}\\ 
\Xhline{3\arrayrulewidth}
Jaccard & 0.718 & 0.734 & 0.756 & 0.755 & 0.756 & 0.777 & 0.770 & 0.773 & 0.779 & 0.772 & \textbf{0.784} \\ 
Sensitivity & 0.836 & 0.859 & 0.839 & 0.847 & 0.830 & 0.847 & \textbf{0.862} & 0.837 & 0.840 & 0.854 & 0.852 \\
Spesificity & 0.935 & 0.914 & 0.960 & 0.953 & 0.957 & 0.965 & 0.951 & \textbf{0.977} & 0.974 & 0.966 & 0.976 \\
Dice & 0.805 & 0.828 & 0.840 & 0.839 & 0.841 & 0.859 & 0.856 & 0.857 & 0.862 & 0.865 & \textbf{0.872} \\
\hline
\end{tabular}}
\end{center}
\label{tab:eval}
\end{table}

\begin{figure}
\centering
\includegraphics[width=12.2cm,height=4.7cm]{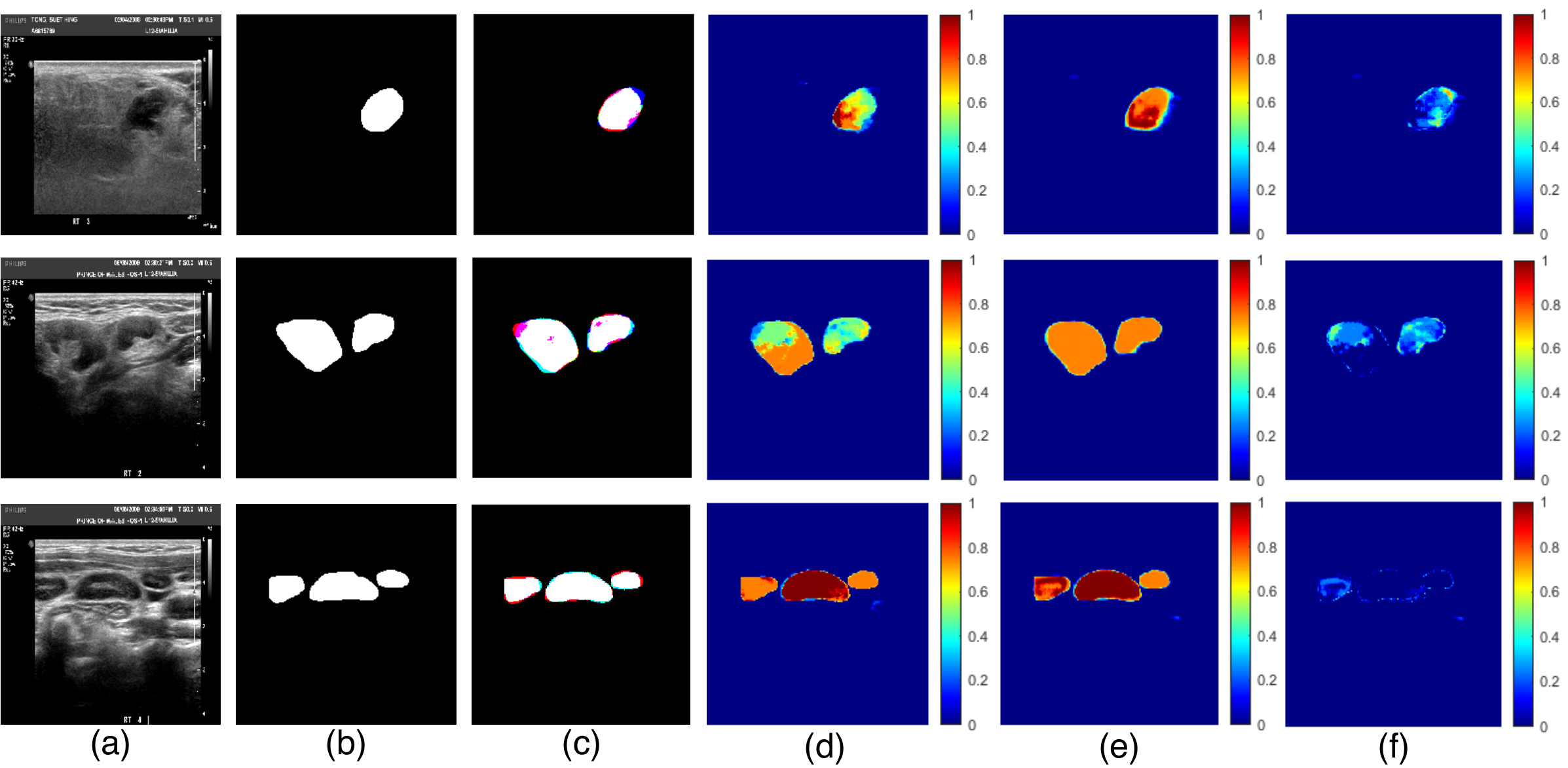}
\caption{Robust feature extraction. (a) A raw ultrasound lymph node image. (b) Ground truth (GT). (c) GT in red, DDN in green, and DDN+ADS in blue. (d)-(f) Segmentation probability for DDN, DDN+ADS, and their difference (f = e - d), respectively.}  
\label{fig:results_ln}
\end{figure}

{
\bibliographystyle{splncs04}
\bibliography{refs}
}